\documentclass[letterpaper, 10 pt, conference]{ieeeconf}  %
\IEEEoverridecommandlockouts
\usepackage{multirow}
\usepackage{cite} 
  \usepackage[pdftex]{graphicx}
  \usepackage[autostyle]{csquotes}
  \usepackage{subcaption}

\usepackage[cmex10]{amsmath} 
\usepackage{amssymb}
\usepackage{booktabs} 

\title{\LARGE \bf
Using Variable Natural Environment Brain-Computer Interface Stimuli for Real-time Humanoid Robot Navigation}

\author{Nik Khadijah Nik Aznan$^{1,3}$, Jason D. Connolly$^{2}$, Noura Al Moubayed$^{1}$ and Toby P. Breckon$^{1,3}$
\thanks{$^{1}$Department of Computer Science,
        Durham University, Durham, UK.}%
\thanks{$^{2}$Department of Psychology,
        Durham University, Durham, UK.}%
\thanks{$^{3}$Department of Engineering,
        Durham University, Durham, UK.}%
}

\begin{document}

\maketitle
\thispagestyle{empty}
\pagestyle{empty}

\begin{abstract}

  This paper addresses the challenge of humanoid robot teleoperation in a natural indoor environment via a Brain-Computer Interface (BCI). We leverage deep Convolutional Neural Network (CNN) based image and signal understanding to facilitate both real-time object detection and dry-Electroencephalography (EEG) based human cortical brain bio-signals decoding. We employ recent advances in dry-EEG technology to stream and collect the cortical waveforms from subjects while they fixate on variable Steady State Visual Evoked Potential (SSVEP) stimuli generated directly from the environment the robot is navigating. To these ends, we propose the use of novel variable BCI stimuli by utilising the real-time video streamed via the on-board robot camera as visual input for SSVEP, where the CNN detected natural scene objects are altered and flickered with differing frequencies (10Hz, 12Hz and 15Hz). These stimuli are not akin to traditional stimuli - as both the dimensions of the flicker regions and their on-screen position changes depending on the scene objects detected. On-screen object selection via such a dry-EEG enabled SSVEP methodology, facilitates the on-line decoding of human cortical brain signals, via a specialised secondary CNN, directly into teleoperation robot commands (approach object, move in a specific direction: right, left or back). This SSVEP decoding model is trained via \textit{a priori} offline experimental data in which very similar visual input is present for all subjects. The resulting classification demonstrates high performance with mean accuracy of 85\% for the real-time robot navigation experiment across multiple test subjects.

\end{abstract}

\IEEEpeerreviewmaketitle

\section{Introduction}

Teleoperation or telepresence is a field within robotics which has been widely utilised for numerous applications. It allows humans to remotely control robots, either whilst being present within the same location, or remotely via the internet \cite{bi2013eeg}. In this work, a humanoid robot is used as teleoperational remote control interface, allowing a human to navigate the robot via the use of Brain-Computer Interface (BCI) based cortical brain bio-signals \cite{Rao2013}. This application can be used widely, for example by severely disabled people as an alternative communication platform with the robot without any actual physical movement \cite{sheng2017design}.

A Brain-Computer Interface is a system that provides a communication and control medium between human cortical signals and external devices \cite{gao2017noninvasive}. One of the primary aims of BCI is to assist or to be used by patients with Complete Locked-In Syndrome in which the end user cannot move or communicate due to paralysis, yet is cognitively intact and can therefore make real, tangible and informed decisions \cite{Rao2013}.

In order to gather the cortical signals from a human test subject, a non-invasive dry-Electroencephalography (EEG) will be used. EEG is a technique where electrodes or sensors are placed on the scalp to capture electrical activity of the brain, without the need to implant them directly into the brain, such as invasive microelectrode arrays \cite{minguillon2017trends}. We utilise the Cognionics Quick-20 dry-EEG Headset that requires no conductive gel and has the additional benefit of being a wireless device; as compared to traditional wet-EEG \cite{Lin2014,mullen2015real}. Wet-EEG requires the cumbersome application of conductive gel, use of an expensive Faraday cage enclosure that prohibits real-world application and scratching of the skin surface via semi-invasive blunted needles in order to lower the impedance values - these measurements represent how usable the connectivity is between the electrodes and the scalp \cite{Lopez-Gordo2014}. It is an alternative approach used to improve the usability of EEG within a BCI context via the elimination of these factors that is required for the wet-EEG approach \cite{lisi2016dry}.

This work explores the creation of a BCI-based application to accurately navigate a humanoid robot in an open environment via the above noted dry-EEG headset. We make use of NAO, a humanoid robot which is equipped with cameras and programmable movement and behavioural features such that different commands can be interpreted to navigate the robot to move toward the object of participant visual and cognitive interest \cite{zhao2017behavior}. To develop this effectively, we employ the features available from the humanoid robot such as streaming the real-time video from the robot as visual input for the SSVEP stimuli. SSVEP is a type of stimulus-evoked neurophysiological response induced simply via subject fixation (or even just via peripheral attention) on visual stimuli and requires almost no \textit{a priori} user training \cite{liu2018research,chiu2017wireless,naz2016recent}. The human cortical signals in the primary visual areas oscillate when visually evoked via these stimuli by a continuously fluctuating sinusoidal cycle \cite{andersen2015driving,mao2017progress}. 

In this paper, we propose a novel variable dry-EEG enabled BCI stimuli for robot navigation utilising a pre-trained object detection convolutional neural network. We perform object detection in real-time derived from the incoming video stream from the robot camera. Our key idea is to make the SSVEP stimuli more natural to the user, as the stimuli (or in our case, objects) will be presented in the context of the real-world scene the robot is currently navigating. Unlike previous stimuli \cite{Lin2014,chen2017single,liu2018research}, in this work the size of each SSVEP flicker region depends on the physical dimensions of the object detected. The detected object pixel regions are flickered at differing on screen frequencies (10, 12, 15 Hz) and the decoded EEG signals are used to navigate the robot to walk toward objects based on the objects selected by the subject (user) via the SSVEP interface.

To perform the dry-EEG signal decoding we use CNN architecture, detailed here \cite{aznan2018classification}, to differentiate between EEG signals by extracting unique features across multiple layers of convolutional transformation optimised over a set of training data \cite{Goodfellow-et-al-2016}. This model is used to classify real-time dry-EEG signals before sending the decoded command to navigate the robot towards the scene object the subject has selected \cite{schirrmeister2017}. 

Following standard practice in the BCI literature, we evaluate the performance of our work by testing the classifier model on real-time humanoid navigation via classification accuracy and Information Transfer Rate (ITR) as performance metrics - the latter representing a quantitative measure of the speed of BCI information transfer \cite{Rao2013}.

In summary, the major contributions of this paper are: 
\begin{itemize}
  \item Use of a novel variable position and size SSVEP BCI stimuli, based on using object detection pixel regions identified in real-time, within the live video stream from a teleoperated humanoid robot traversing a real-world natural environment. 
  
  \item An offline dry-EEG enabled SSVEP BCI signal decoding (classification) result achieving mean accuracy of 84\% with the use of variable stimuli size and on-screen stimuli positioning (the first such study to accomplish this).
  
  \item Demonstrable real-time BCI teleoperation of a humanoid robot, based on the use of naturally occurring in-scene stimuli, with a peak mean accuracy of 85\% and ITR of 15.2 bits per minute (bpm) when evaluated over multiple test subjects (teleoperation users).
\end{itemize}

\section{Related Work}

There have been many prior studies utilising humanoid robots with EEG signals for various BCI applications \cite{bi2013eeg, gao2017noninvasive, mao2017progress}. In this section we will focus on the studies making use of SSVEP within this context.


The work of \cite{zhao2017behavior} proposed the use of behaviour-based SSVEP to control a telepresence humanoid robot to walk in a cluttered environment, with the tasking of approaching and picking up a target. They controlled the robot by classifying 4 sets of movements with a total of fourteen behaviours of the robot. One visual stimuli is used to select the behaviour set and the remainder are used to encode the behaviours. The user interface of the system consists of five fixed stimuli symbols (five frequency values), a display for a live video feedback and a display for the current posture of the robot. The task completed with an average success rate of 88$\%$, an average response time of 3.48 s and an average ITR of 27.3 bpm.

Similar research has been carried out using SSVEP stimuli to control robot-like behaviour in \cite{liu2018research} and \cite{chen2017single} in which these authors again used fixed size and position stimuli symbols with differing frequencies that indicate different directions for the robot to move toward. In \cite{liu2018research} the authors controlled a mobile robot by using 3 different SSVEP frequencies by moving forward or turning to the left or right in order to avoid the obstacles. The stimuli in \cite{chen2017single} consisted of four fixed flickering boxes where each frequency was used to command a mobile robot (forward, backward, turn counter-clockwise/clockwise) to navigate the robot through a maze path.

There are two notable studies that have integrated object detection and recognition \cite{sheng2017design, tidoni2017role}. In \cite{sheng2017design}, the authors used seven different frequencies to navigate a mobile robot to a storage rack to grasp an object and delivered into a dustbin with an average mean accuracy of 89.4\%. The approach employed an AdaBoost algorithm with Haar features to recognise three objects on the rack for subjects to choose. However, the recognised objects were not flickered as stimuli - instead, there were separate fixed stimuli designed with three different frequencies corresponding to each object.

The authors in \cite{tidoni2017role} used SSVEP with a hybrid-mask feature in which a 3D textured model was rendered and flickered on certain scene objects. In this case, three similar cans which are recognised offline. Subjects for this study teleoperated a humanoid robot HRP-2 (located in Japan from Italy) to control the robot to firstly, grasp a can from a table, navigated the robot to a second table where the robot is required to drop the can on a marked target.

In this present work, by taking advantage of the on-board camera on our humanoid robot and the high-performance scene object detection model of \cite{liu2016ssd}, we instead use variable BCI stimuli, embedded within the scene video feed. This is achieved by flickering the flexible size detected object pixel regions with differing SSVEP frequencies. This occurs in the real-time as the humanoid robot navigates a natural indoor environment. In contrast to earlier work \cite{zhao2017behavior, liu2018research, chen2017single, sheng2017design, tidoni2017role}, our stimuli vary both in terms of pixel pattern, size and on-screen position in-conjunction with the changing nature of the environment the robot is navigating through.

\begin{figure}[!h]
  \centering
  \includegraphics[width=1.0\linewidth]{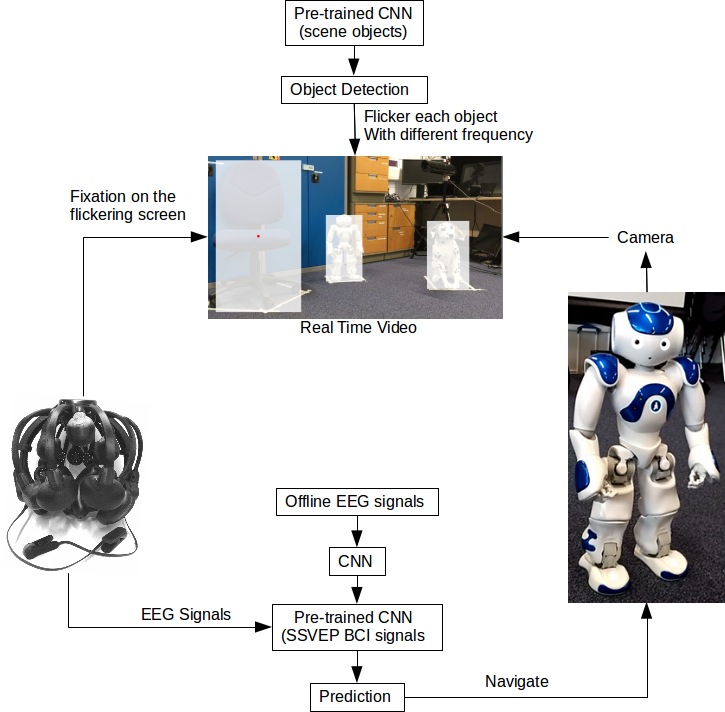}
  \caption{Overview of the experimental approach proposed.}
  \vskip -20pt 
  \label{fig_sim} 
  \end{figure}

  \begin{figure*}[!h]
    \centering
    \vskip 5pt
    \begin{subfigure}[b]{0.32\textwidth}
      \includegraphics[width=\linewidth]{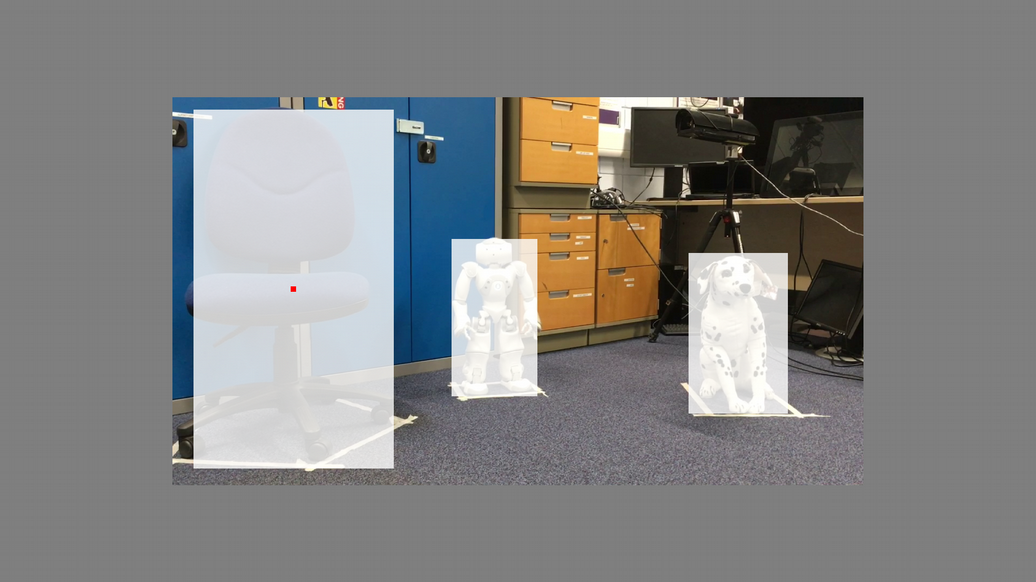}
        \caption{}
        \label{fig:Stimuli01}
    \end{subfigure}
    \begin{subfigure}[b]{0.32\textwidth}
      \includegraphics[width=\linewidth]{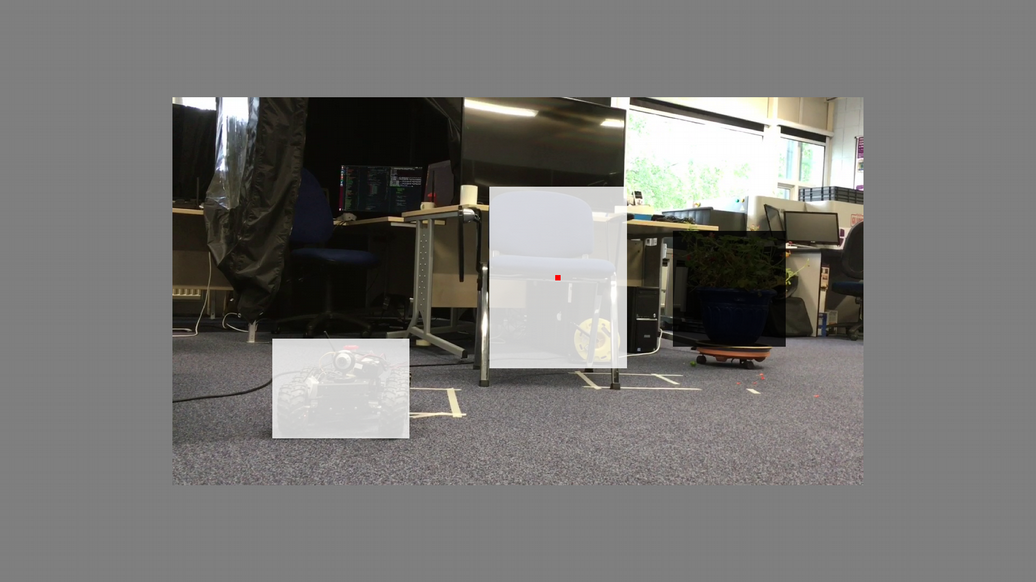}
        \caption{}
        \label{fig:Stimuli02}
    \end{subfigure}
    \begin{subfigure}[b]{0.32\textwidth}
      \includegraphics[width=\linewidth]{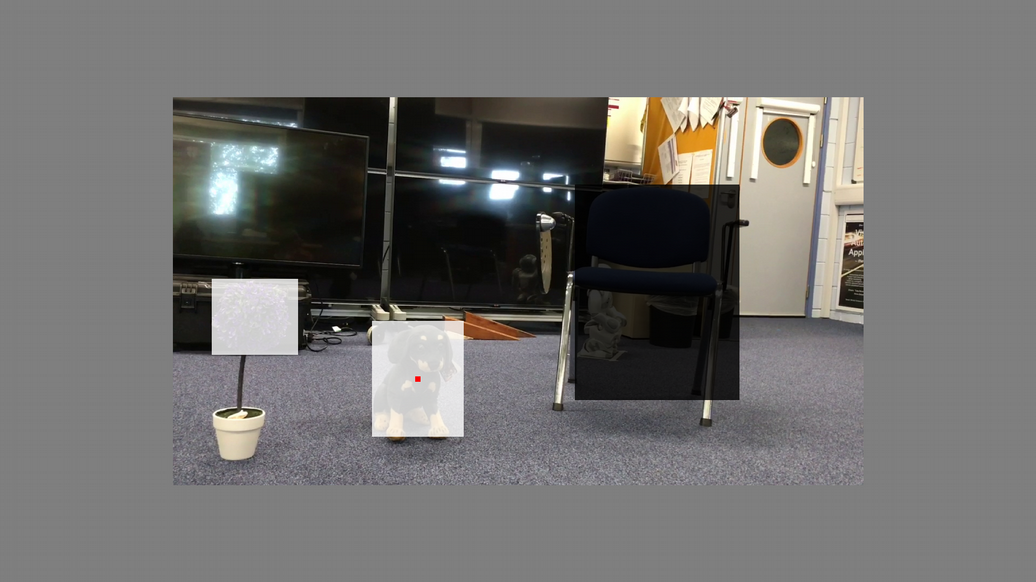}
        \caption{}
        \label{fig:Stimuli03}
    \end{subfigure}
    \vskip -15pt
    \begin{subfigure}[b]{0.32\textwidth}
      \includegraphics[width=\linewidth]{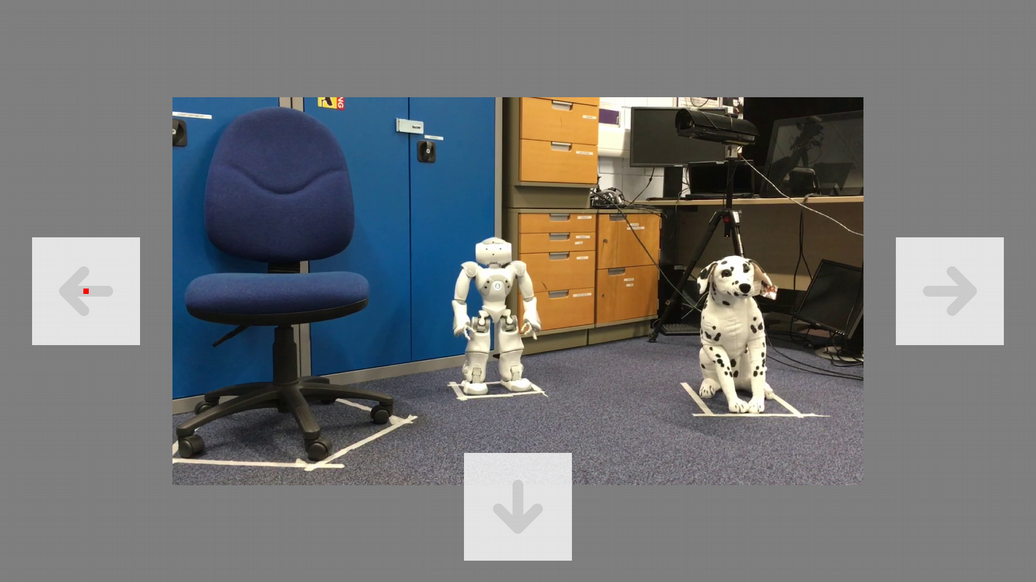}
        \label{fig:Stimuli04}
    \end{subfigure}
    \begin{subfigure}[b]{0.32\textwidth}
      \includegraphics[width=\linewidth]{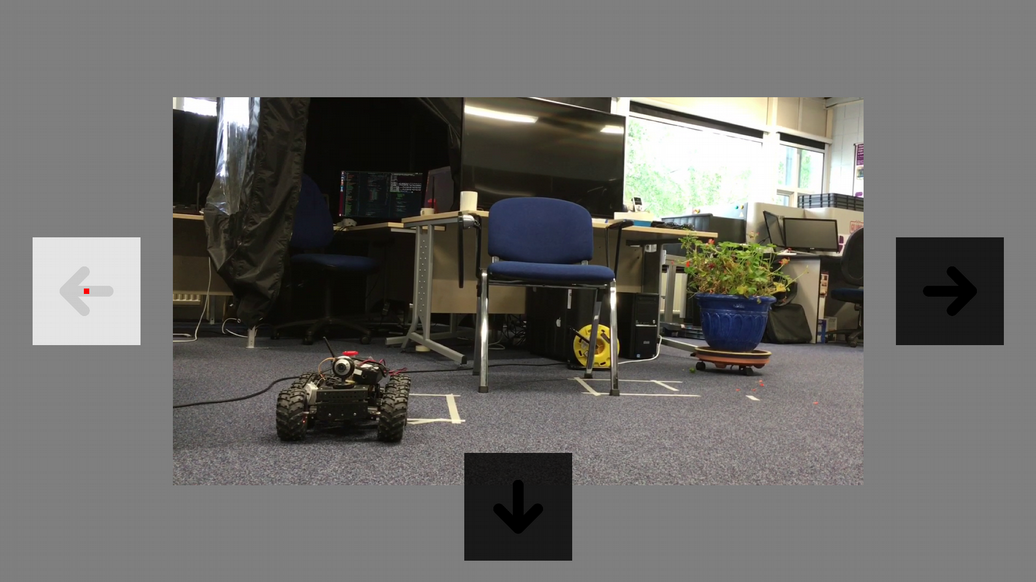}
        \label{fig:Stimuli05}
    \end{subfigure}
    \begin{subfigure}[b]{0.32\textwidth}
      \includegraphics[width=\linewidth]{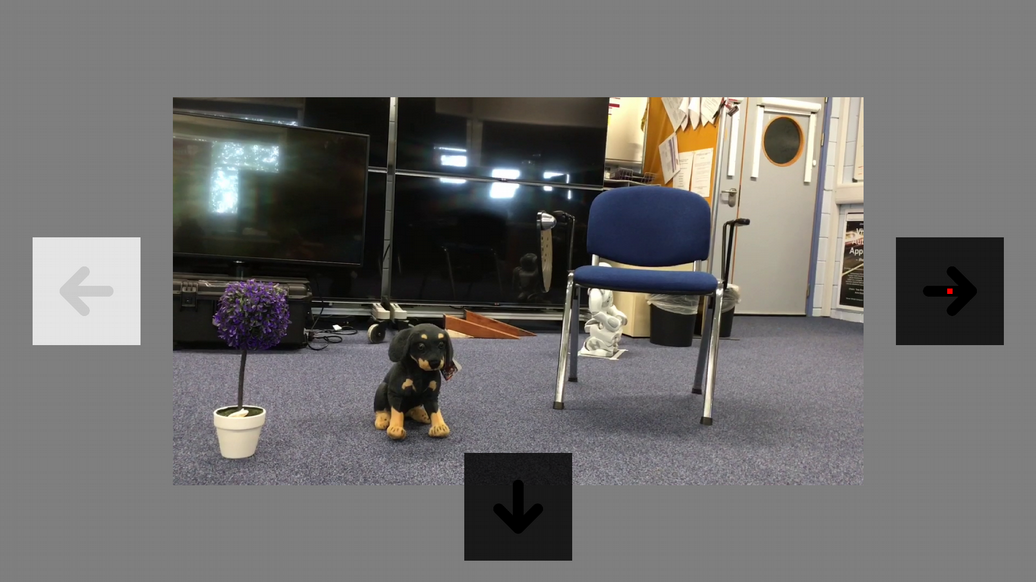}
        \label{fig:Stimuli06}
    \end{subfigure}
    \vskip -10pt
    \caption{Variable SSVEP stimuli based on object detection flickering (top row) to navigate the robot towards the object and navigational arrow flickering (bottom row) to move the robot facing a new environmental scene.}
    \label{fig:Stimuli}
    \vskip -15pt
  \end{figure*}
  
\section{Methodology}

In this section, we present the four primary experimental components; variable BCI stimuli, streaming dry-EEG signals, EEG signal classification and robot navigation. The overall setup and data flow of the experiment is shown in Figure \ref{fig_sim}.

We use the on-board camera to stream video from the natural environment to a monitor display in front of the BCI subject (user). Using the CNN-based object detection model of \cite{liu2016ssd}, detected scene objects are identified and flickered with a unique on-screen SSVEP frequency (from set: 10Hz, 12Hz, 15Hz). EEG signals from the subject are streamed using the dry-EEG headset whilst they fixate on a flickering on-screen scene object. A CNN pre-trained on an \textit{a priori} offline dataset is then used to infer the class of the EEG signals in real-time. This prediction is used to navigate the robot towards the corresponding scene object the subject is fixated upon.

\subsection{Variable BCI Stimuli}
\label{sec:exerimentalsetup} 

In order to translate the cortical signals, we use SSVEP as the neurophysiological brain response for subjects. The stimuli are embedded into the real-time video streaming from the on-board robot camera (RGB colour, resolution: 1280$\times$960). Based on pre-trained object detection, we flicker the on-screen display frequency of objects by rendering black/white polygon boxes on top of the objects with display frequency modulations of 10, 12 and 15 Hz \cite{mao2017progress}.

In this work, we employ the pre-trained Single Shot MultiBox (SSD) Object Detector CNN \cite{liu2016ssd}. This CNN was trained by using the 12 objects class from the COCO dataset \cite{lin2014microsoft}. We present the stimuli using\cite{peirce2007psychopy} on a 60Hz refresh rate LCD monitor.

The teleoperation interface display alternates between this detected object flickering and navigational arrow flickering one after another as illustrated in Figure \ref{fig:Stimuli}. The additional use of the navigational arrow stimuli enables the subject to navigate the robot when there is no new object detected within the scene, for example, when the robot is too close to the previously subject (user) selected object. 

\subsection{Dry-EEG Signal Streaming}
\label{sec:classification}

\begin{figure}[!h]
  \vskip -10pt
  \centering
  \includegraphics[width=0.75\linewidth]{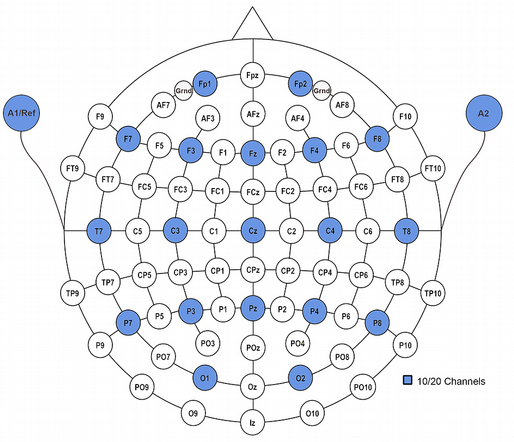} 
  \caption{The location of the electrodes of the dry-EEG headset highlighted in blue within the 10/20 EEG reference model.}
  \vskip -5pt
  \label{fig:sensors}
  \end{figure}

We use the Cognionics Quick-20 20-channel dry-EEG headset to stream the cortical signals from three healthy subjects (S01, S02, S03) whilst each subject is sat in front the variable SSVEP stimuli. The dry-EEG headset provides 19 channels and A2, reference and ground as in Figure \ref{fig:sensors} with a 10-20 compliant sensor layout (international standard for reproducible sensor placement across different EEG experiments \cite{mullen2015real}).

This portable and wireless headset is straightforward and easy-to-use as it does not require neither any skin preparation prior to use or conductive gel (as wet-EEG).

During the experiments, we stream the signals over nine sensors; parietal and occipital cortex (P7, P3, Pz, P4, P8, O1 and O2) \cite{gao2017noninvasive,Lin2014,mao2017progress}, frontal centre (Fz) and A2 reference at 500 Hz sampling rate for three seconds per trial. The odd numbers represent the left hemisphere of the brain, whilst the even represent the right hemisphere (Figure \ref{fig:sensors}).

The dry-EEG headset requires proprietary data acquisition software, used to measure impedance values before use to ensure optimal-quality dry-EEG signals. In addition, it streams the data from the headset to a computer and allows networked data access to send the data streaming over the network (between two different computers, for example).

\subsection{EEG Signals Classification}
\label{sec:cnn}

To decode the dry-EEG signals efficiently in order to ensure effective teleoperation of the robot, we use our deep CNN architecture of \cite{aznan2018classification}) (see reference for more details) for signal to object/motion label classification. 

During the offline experiments, subjects fixate to one of the flickering stimuli. The cortical brain signals from each subject are collected for 40 experimental trials per SSVEP class to form the offline \textit{a priori} training sets or training the CNN model per subject (offline calibration).

\begin{figure}[!h]
  \centering
  \includegraphics[width=0.85\linewidth]{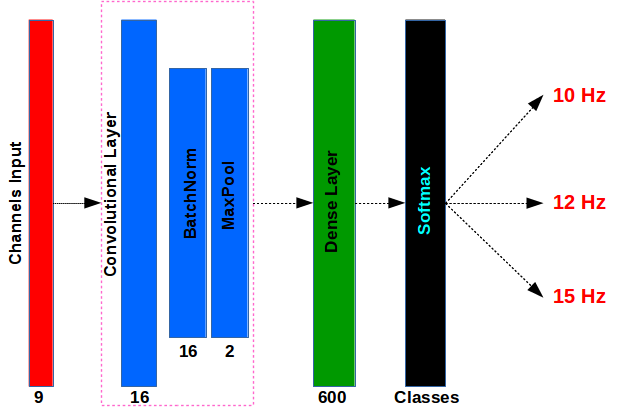} 
  \caption{The 1D CNN architecture used to classify the EEG signals for both offline dataset and real-time experiments (SCU, highlighted in pink).}
  \vskip -5pt
  \label{fig:CNN}
\end{figure}

We train a SSVEP Convolutional Unit (SCU) CNN architecture \cite{aznan2018classification}, comprising of a 1D convolutional layer, batch normalization and max pooling (as detailed in Figure \ref{fig:CNN}) by using the offline priori experimental datasets. We first bandpass filter the incoming sigmals between 9 to 100 Hz in order to reduce undesired high or low frequencies that are not of interest in this work. The filtered signals which consist of nine input channels are transformed by using a large initial convolutional filter to capture the frequencies we are interested in classifying the dry-EEG data. The SCU CNN model is trained using backpropagation with stochastic gradient descent \cite{lecun2015deep}.

For this training, the key hyperparameters, initially chosen via a grid-search over a validation set, are L2 weight decay scaling 0.004, dropout level 0.5, convolution kernel size 1$\times$10, kernel stride 4, maxpool kernel size 2, categorical cross entropy as the optimisation function, ADAM gradient descent algorithm \cite{kingma2014adam} and ReLU as the activation function on all hidden layers.

\subsection{Robot Navigation}
\label{sec:Navigation}

The experiment begins with the robot facing a scene containing objects which are detected to generate on-screen SSVEP stimuli pixel regions as previously outlined. The subject (teleoperation user) fixates on one particular object from which robot navigation is performed using the high level mobility functions of the NAO humanoid robot platform (Figure \ref{fig:Stimuli}), based on the decoding of the corresponding SSVEP signals by the pre-trained SCU CNN model (Section \ref{sec:cnn}).

Once these BCI signals are classified as a selected scene object by the subject (user), we then calculate the required robot motion trajectory. As we cannot acquire depth information directly from the monocular camera on the robot, we acquire the distance and the angle of view of the chosen object following the photogrammetric approach of \cite{kundegorski2014photogrammetric}. As such, the distance of the object $Z$ can be calculated as:

\begin{equation}
  \label{eq:distance}
 Z = f'{\frac{Y}{y}},
\end{equation}

where $Z$ is the distance in metres, $f'$ is the focal length (pixels), $Y$ is the object height (metres) and $y$ is the object height in the image (pixels) $f'$ as:

\begin{equation}
  \label{eq:focalpx}
  f' = H{\frac{f}{h}},
\end{equation}

where $H$ is height of the image (pixels), $f$ is the focal length in metres and $h$ is the sensor height in metres.

The angle of view $(AoV)$ of the object from the camera in radian based on the horizontal position x of the image in pixel can be calculated as follows:

\begin{equation}
  \label{eq:angle}
  AoV = \frac{x}{f'}.
\end{equation}

\begin{figure}[!h]
  \centering
  \includegraphics[width=0.75\linewidth]{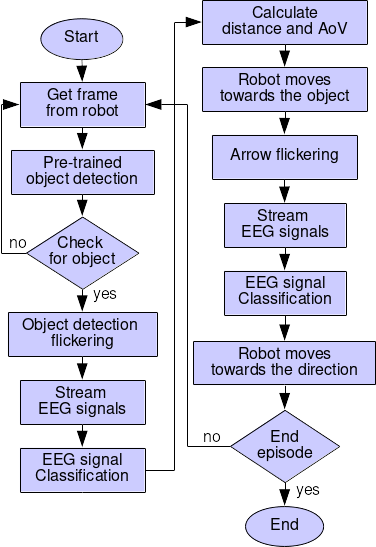} 
  \caption{Flowchart of real-time robot navigation.}
  \vskip -5pt
  \label{fig:RT}
  \end{figure}

When the robot navigates within a given distance and angle trajectory of the subject-selected object, the BCI on-screen interface display alternates to the navigational arrow display (left, right, backwards \textendash{} Figure \ref{fig:Stimuli}) using the specific SSVEP frequencies of 10, 12 and 15 Hz. These frequencies intend to facilitate robot motion at 90 degree turns left/right or a 180 degree about turn. Subjects similarly attend to one of these SSVEP stimuli which, once decoded by the SCU CNN model, facilitate general robot motion in the environment until further scene objects are detected within the scene traversal. A flow diagram operation of the real-time experimental teleoperation of the NAO robot through the environment in this alternating object-stimuli and navigational-stimuli manner is presented in Figure \ref{fig:RT}.

\begin{figure}[!h]
  \centering
  \includegraphics[width=1.0\linewidth]{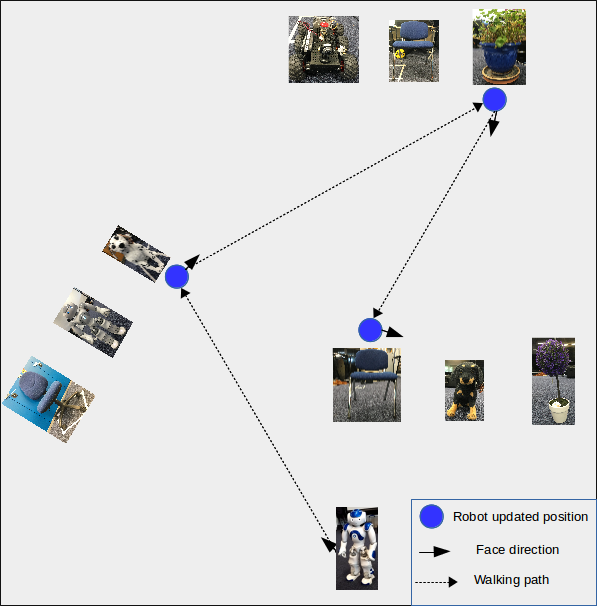} 
  \caption{Navigation plan for real-time experimentation.}
  \vskip -5pt
  \label{fig:plan} 
\end{figure}

The experimental navigation plan used during the real-time experiments presented in this study is shown in Figure \ref{fig:plan}. Under these conditions, we repeat the experimental episode five times per subject to demonstrate the repeatability and robustness of our approach.

\section{Results and Discussion}

In this section, we present the results from the offline classification and the real-time experiment classification using the metrics of classification accuracy and Information Transfer Rate (ITR) in bits per minute (bpm).

\subsection{Offline Statistical Performance} 
\label{sec:Offline}

The result for the classification accuracy and the ITR of the offline experiment are presented in Table \ref{table:offline result}. ITR is the speed of BCI in term of bit rate transfer which is the amount of the information throughput by a system per minute \cite{Rao2013}.

\begin{table}[h!]
  \vskip 5pt
  \begin{center}
  \begin{tabular}{c c c c c c}
  \toprule
  \textbf{Subject} & \textbf{S01} & \textbf{S02} & \textbf{S03}\\
  \midrule \midrule
  \textbf{Accuracy} & 0.96$\pm$0.02 & 0.80$\pm$0.09 & 0.75$\pm$0.12 \\
  \textbf{ITR (bpm)} & 23.90$\pm$0.72 & 12.16$\pm$0.95 & 9.61$\pm$0.89 \\
  \bottomrule 
  \end{tabular}
  \end{center}
  \vskip -5pt
  \caption{Mean accuracy and ITR with standard deviation for offline classification over 10-fold cross validation.}
  \vskip -15pt
  \label{table:offline result}
  \end{table}

  ITR is a suitable BCI performance metric, as a high ITR is dependent upon high accuracy. The ITR is calculated as in \cite{tidoni2017role}:

\begin{equation}
  \label{eq:ITR}
 ITR = {\frac{B}{T}},
\end{equation}

where T is the time taken to classify a trial in minutes and B is the bits per trial:

\begin{equation}
  \label{eq:B-ITR}
 B = {log_2}(N)+P{log_2}(P)+(1-P){log_2}({\frac{1-P}{N-1}}), 
\end{equation}

where N is the number of possible selections (N = 3) and P is the correct selection accuracy.

For the offline experiment, the time taken is based on the total flickering time per trial (3 seconds) plus the average of time the classifier takes to train and classify a trial. The data collected during the offline experimental phase is used to train the model for real-time experimentation. However, in order to demonstrate statistical performance of our SCU CNN architecture on this task, we present mean accuracy over 10-fold cross validation per subject. This is used as the P value to calculate the value for B (Equation \ref{eq:B-ITR}).
  
\subsection{On-line Real-time Performance}

\begin{figure*}[!h]
  \vskip 5pt
  \centering
  \begin{subfigure}[b]{0.29\textwidth}
    \includegraphics[width=1.0\linewidth]{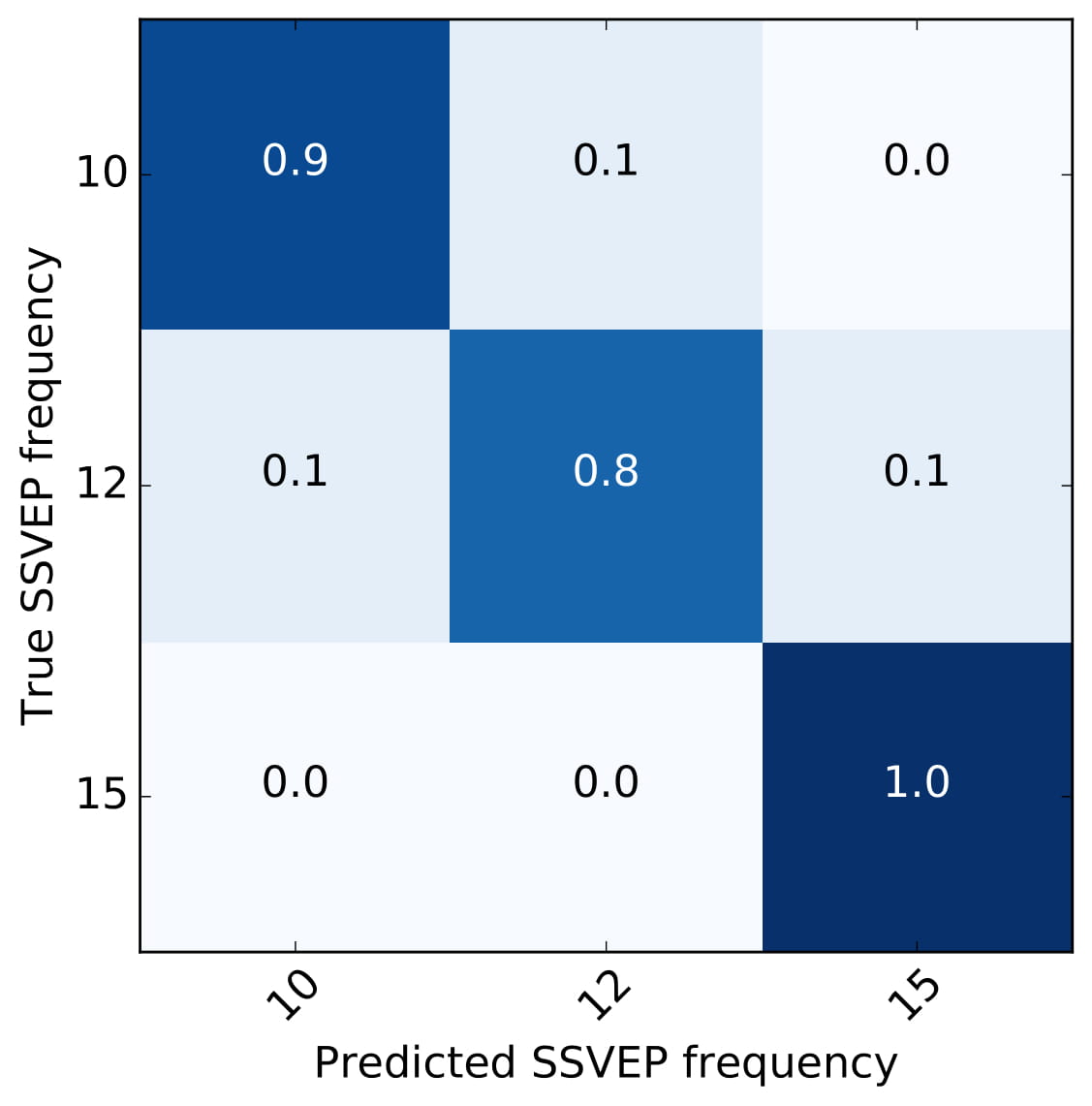}
      \caption{Subject S01}
      \label{fig:S01}
  \end{subfigure}
  \hfill
  \begin{subfigure}[b]{0.29\textwidth}
      \includegraphics[width=1.0\linewidth]{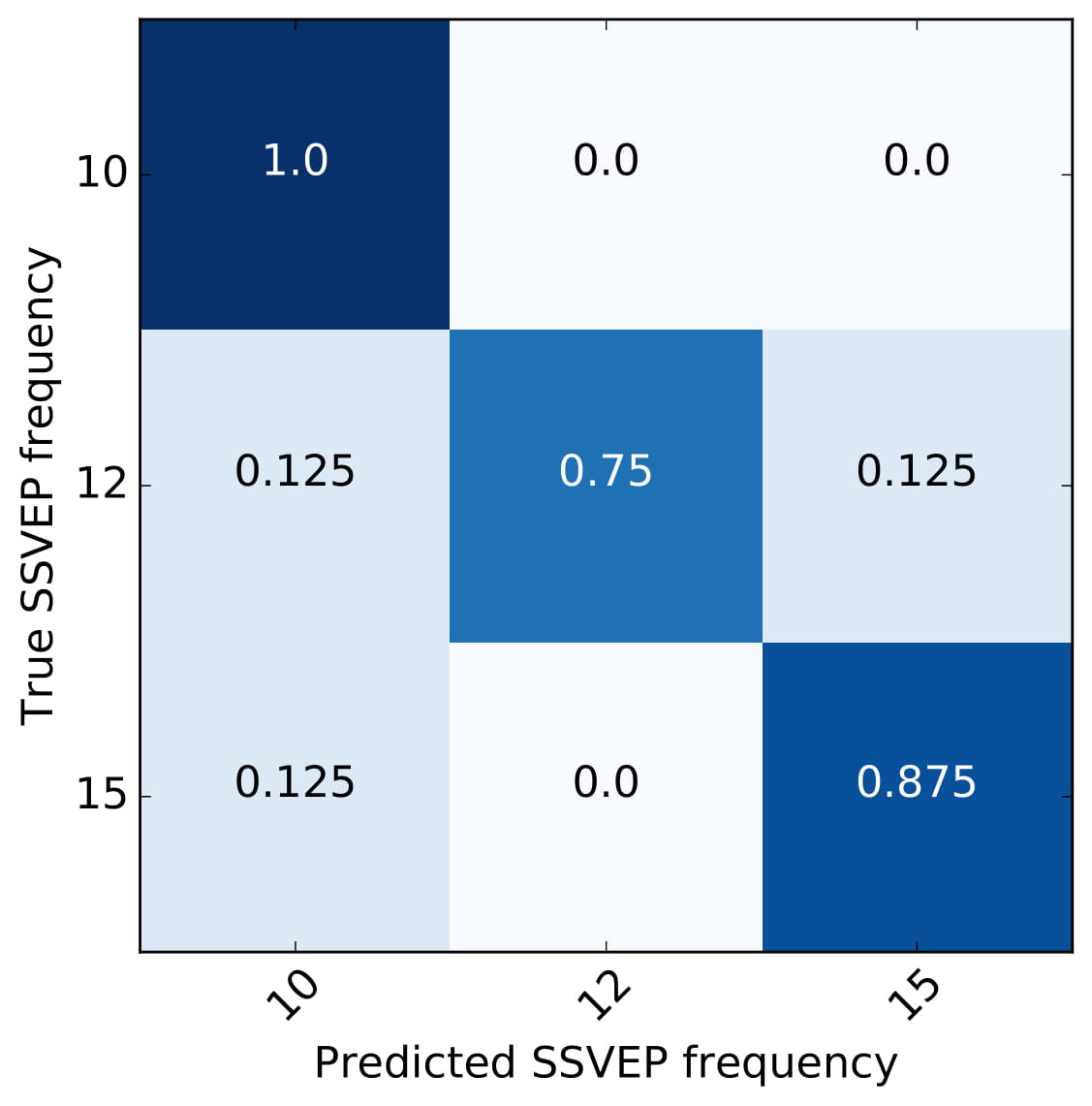}
      \caption{Subject S02}
    \label{fig:S02}
  \end{subfigure}
  \hfill
  \begin{subfigure}[b]{0.29\textwidth}
      \includegraphics[width=1.0\linewidth]{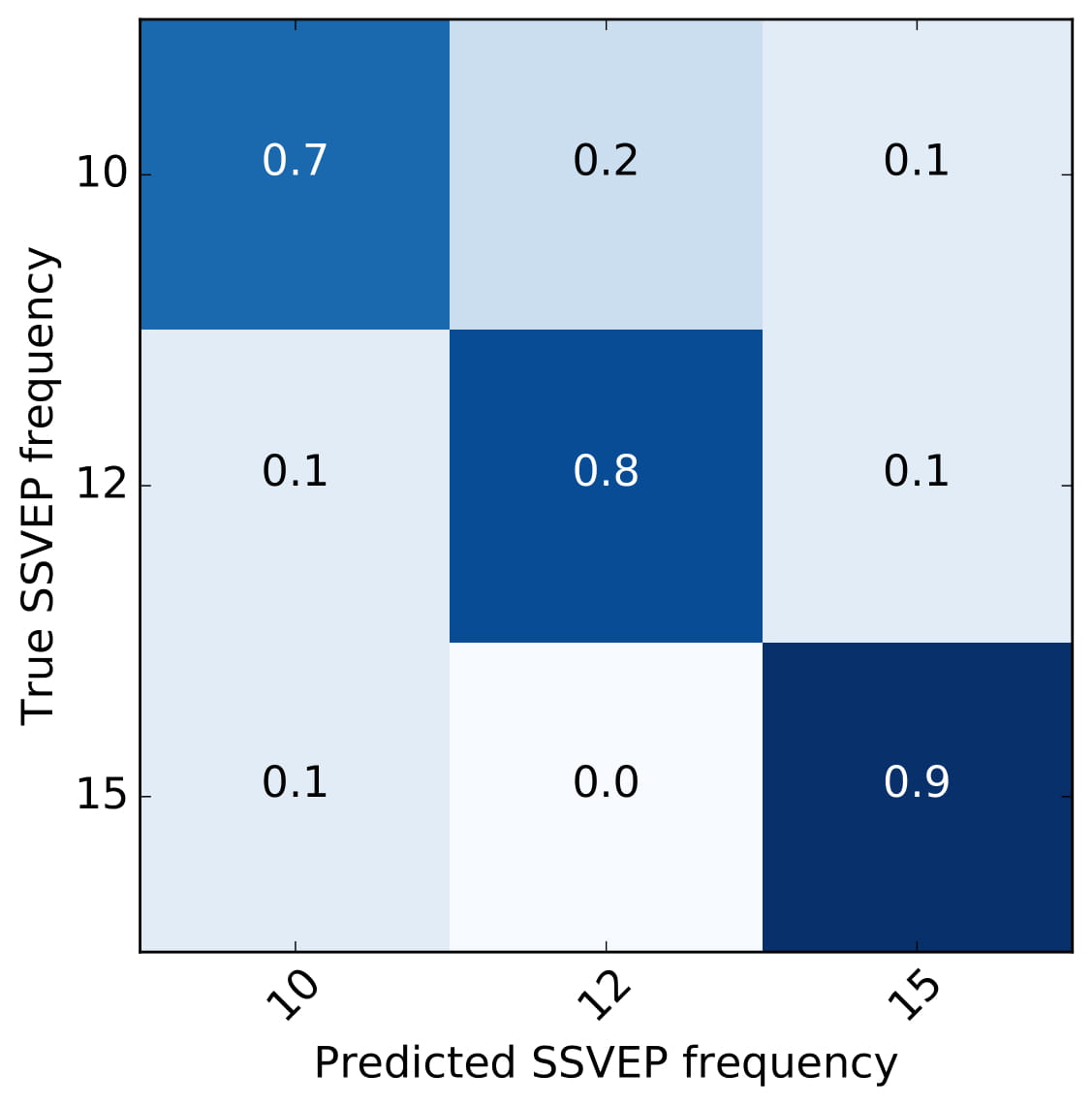}
      \caption{Subject S03}
    \label{fig:S03}
  \end{subfigure}
  \caption{Confusion matrices for the classification of real-time EEG signals during the robot navigation (maximal result being \textit{accuracy} = 1.0 in the matrix diagonals). }
  \label{fig:cm}
  \vskip -10pt
\end{figure*} 

The results of the on-line experimental phase are presented in Table \ref{table:real-time result} where we can see the correlation between the results from both experiments. Overall, the results demonstrate extremely high accuracy for all of the subjects tested. 

\begin{table}[h!]
  \begin{center}
  \begin{tabular}{c c c c c c}
  \toprule
  \textbf{Subject} & \textbf{S01} & \textbf{S02} & \textbf{S03}\\
  \midrule \midrule
  \textbf{Experiment 1} & 1.00 & 1.00 & 1.00 \\
  \textbf{Experiment 2} & 1.00 & 0.83 & 0.50 \\
  \textbf{Experiment 3} & 0.83 & 0.83 & 0.83 \\
  \textbf{Experiment 4} & 0.83 & 1.00 & 1.00 \\
  \textbf{Experiment 5} & 0.83 & 0.67 & 0.67 \\
  \textbf{Mean Accuracy} & 0.90$\pm$0.08 & 0.87$\pm$0.12 & 0.80$\pm$0.19 \\
  \textbf{Mean ITR (bpm)} & 16.8$\pm$0.10 & 15.6$\pm$0.12 & 13.2$\pm$0.16 \\
  \bottomrule
  \end{tabular}
  \end{center}
  \vskip -5pt
  \caption{Accuracy for each experiment and mean accuracy and ITR with standard deviation for real-time classification.}
  \vskip -10pt
  \label{table:real-time result}
  \end{table}



Our results demonstrate a strong statistical performance, with a mean accuracy of 0.85 across all subjects. This is comparable to \cite{zhao2017behavior} which obtained 0.88 accuracy, despite our work using a variable SSVEP stimuli. As ITR represents the speed of the real-time information transfer from stimuli to motion command generation, the time taken is measured from the beginning of a stimuli flashing until getting a prediction. We can thus improve the ITR further via reducing the flickering time during the real-time experiment.

Figure \ref{fig:cm} represents per-class confusion matrices for the real-time classification and highlights overall good accuracy across all classes for all three subjects (users), although the middle class (12 Hz) is more difficult to classify than the rest of the classes.

\begin{figure}[!h]
  \centering
  \begin{subfigure}[b]{0.23\textwidth}
    \includegraphics[width=\linewidth]{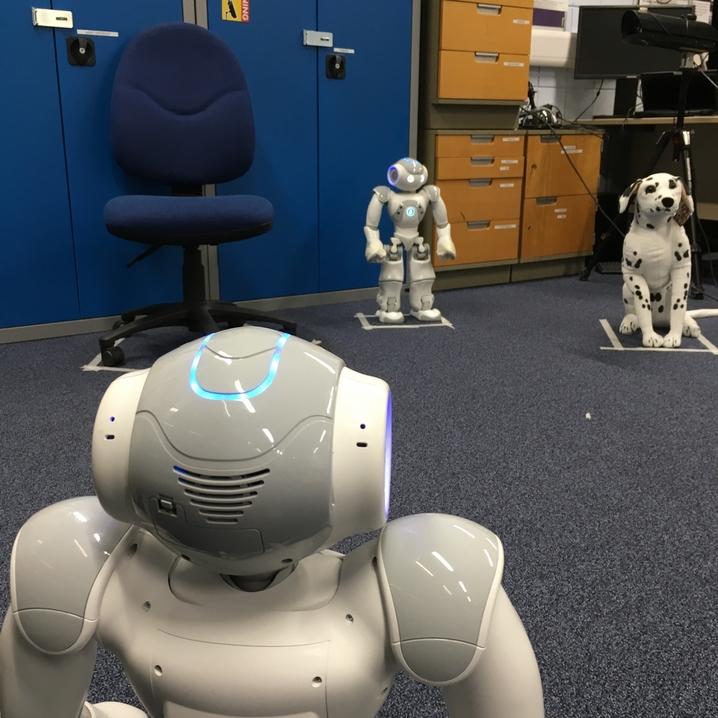}
      \label{fig:E01}
  \end{subfigure}
  \begin{subfigure}[b]{0.23\textwidth}
    \includegraphics[width=\linewidth]{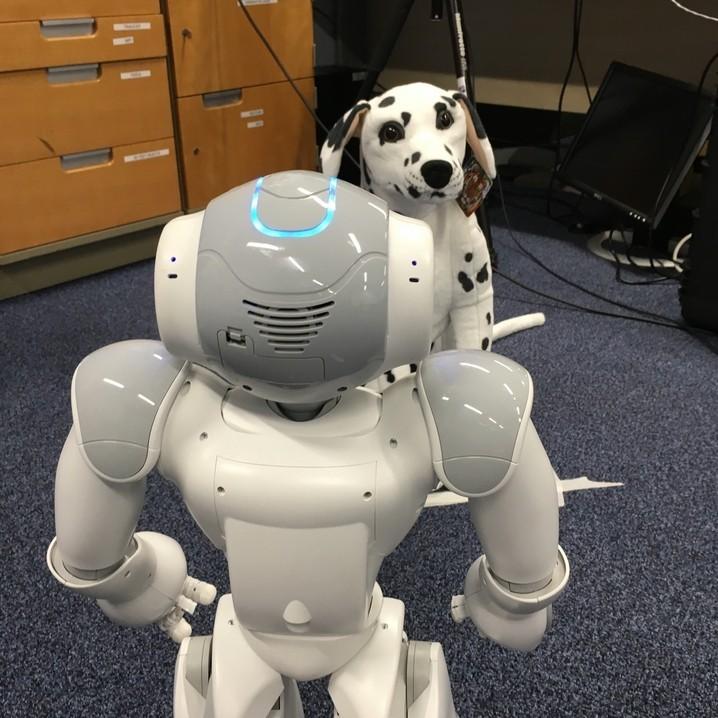}
      \label{fig:E02}
  \end{subfigure}
  \vskip -10pt

  \begin{subfigure}[b]{0.23\textwidth}
    \includegraphics[width=\linewidth]{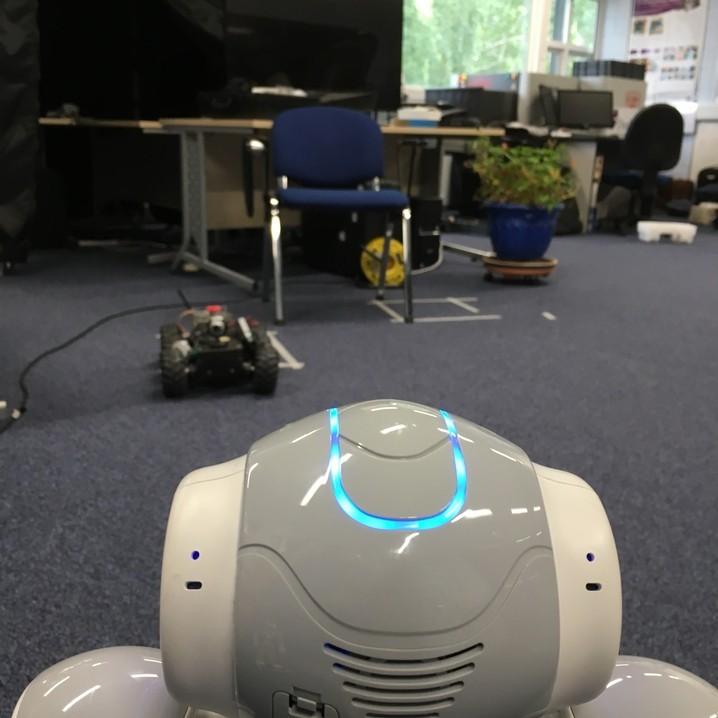}
      \label{fig:E03}
  \end{subfigure}
  \begin{subfigure}[b]{0.23\textwidth}
    \includegraphics[width=\linewidth]{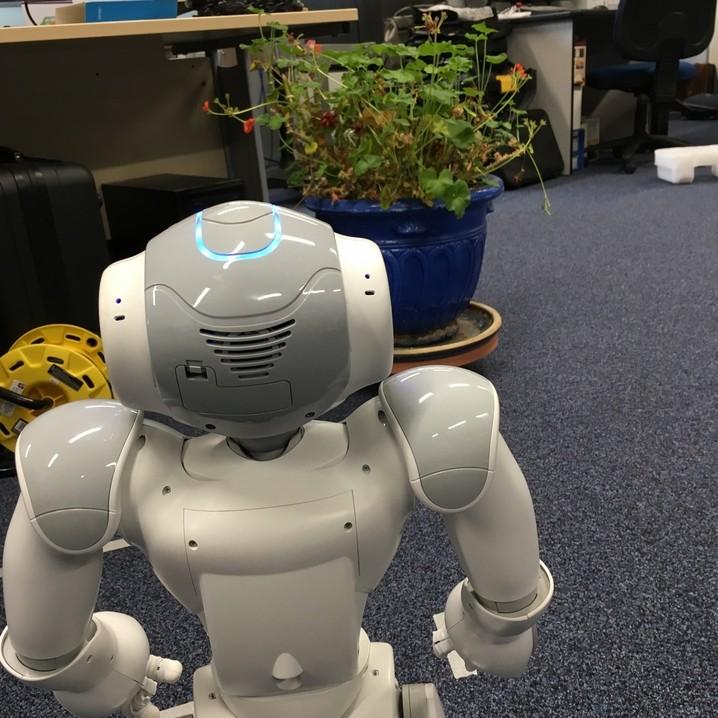}
      \label{fig:E04}
  \end{subfigure}
  \vskip -10pt

  \begin{subfigure}[b]{0.23\textwidth}
    \includegraphics[width=\linewidth]{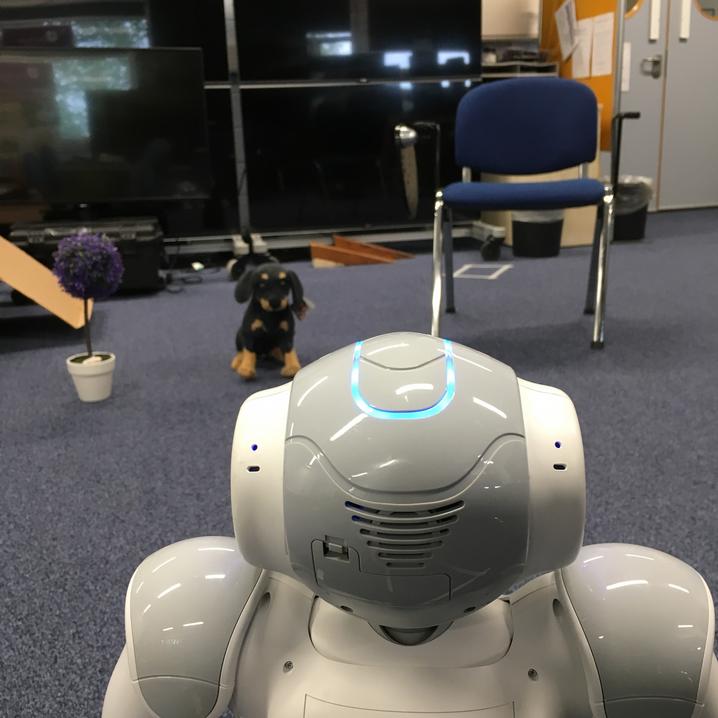}
      \label{fig:E05}
  \end{subfigure}
  \begin{subfigure}[b]{0.23\textwidth}
    \includegraphics[width=\linewidth]{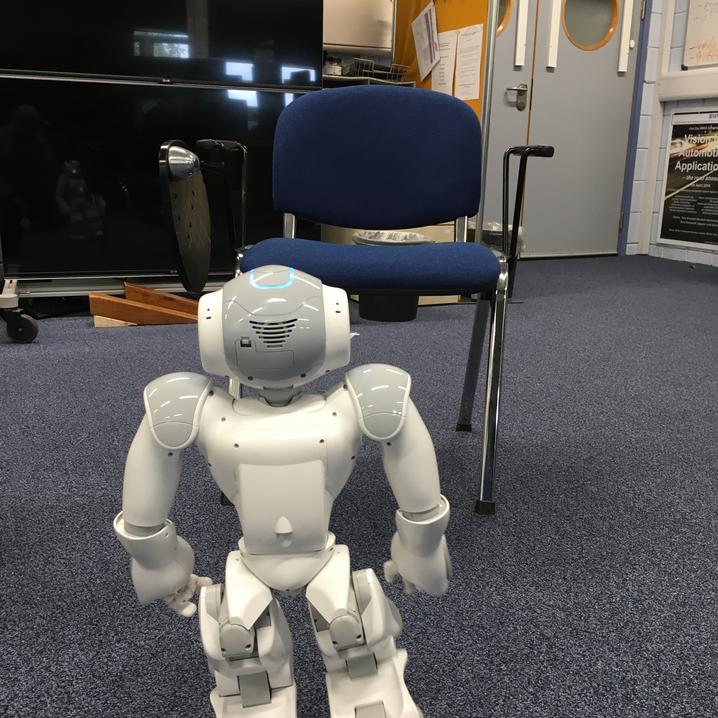}
      \label{fig:E06}
  \end{subfigure}
  \vskip -10pt
  \caption{Sample of humanoid robot navigation during a real-time experiment.}
  \label{fig:experiment}
  \vskip -15pt
\end{figure}

Figure \ref{fig:experiment} illustrates the real-time experimental environment such as the view from the robot and the robot approaching an object. The angle of direction from the robot to the selected objects can vary from one experiment to another, because the calculation of distance and direction is based on the bounding box from the object detection and the angle of view of an object on the plane. The detected bounding box for the scene object can vary and the angle of view of an object can change with the slightest movement of either the robot or the robot head (where the camera is located).

\section{Conclusion}

In this work, we present a number of novel contributions spanning the use of variable SSVEP stimuli (pattern, size, shape) as an enabler to future telepresence BCI applications in a real-world natural environment. We integrate recent advances in the use of deep CNN architectures for both scene object detection and dry-EEG bio-signal decoding. Within this context, we develop a novel SSVEP interface to flicker the on-screen frequency of naturally occurring objects detected within the scene, as seen from the on-board camera of a teleoperated robot, and decode these dry-EEG brain-based bio-signals based on the frequency of the visual fixation detected to navigate the robot within the scene. Uniquely, we train and utilize a common CNN model (SCU, Figure \ref{fig:CNN}) for use with SSVEP stimuli that vary in size, on-screen position and internal (pixel pattern) throughout the duration of the experiment, significantly advancing such decoding generality against prior work in the field \cite{sheng2017design, tidoni2017role}. Our evaluation is presented in terms of accuracy and ITR, both on the \textit{a priori} experimental training set used for the off-line training phase (via cross validation) and the on-line real-time teleoperated navigation of a humanoid robot through a natural indoor environment. The introduction of these highly novel and variable BCI SSVEP stimuli, based on scene object occurrence, demonstrates adaptable BCI-driven robot teleoperation within a natural environment (without scene  markers  and  alike). Strong statistical classification performance is observed, comparable to and often exceeding those reported in the general BCI literature \cite{zhao2017behavior}, despite the introduction of the serious challenges associated with variable SSVEP stimuli. 

Future work will look to improve generalisation performance over additional test subjects, increase both scene complexity and teleoperative duration as well as considering aspects of robot interaction within the environment.




\bibliographystyle{IEEEtran}
\bibliography{ref}

\end{document}